\documentclass[10pt,twocolumn,letterpaper]{article}

\usepackage[pagenumbers]{cvpr} 

\usepackage{graphicx}
\usepackage{amsmath}
\usepackage{amssymb}
\usepackage{booktabs}

\usepackage[pagebackref,breaklinks,colorlinks]{hyperref}

\usepackage[capitalize]{cleveref}
\crefname{section}{Sec.}{Secs.}
\Crefname{section}{Section}{Sections}
\Crefname{table}{Table}{Tables}
\crefname{table}{Tab.}{Tabs.}

\def\cvprPaperID{9761} 
\def\confName{CVPR}
\def\confYear{2022}

\begin{document}

\title{Segmentation-Reconstruction-Guided Facial Image De-occlusion}

\author{
Xiangnan Yin\textsuperscript{\rm 1}, Di Huang\textsuperscript{\rm 2}, Zehua Fu\textsuperscript{\rm 2}, Yunhong Wang\textsuperscript{\rm 2}, Liming Chen\textsuperscript{\rm 1}\\
\textsuperscript{\rm 1} Ecole Centrale de Lyon
\textsuperscript{\rm 2} Beihang University\\
\tt\small{\{yin.xiangnan,liming.chen\}@ec-lyon.fr}, \tt\small{\{dhuang,yhwang\}@buaa.edu.cn}, \tt\small{zehua\_fu@163.com} 
}
\maketitle

\begin{abstract}
Occlusions are very common in face images in the wild, leading to the degraded performance of face-related tasks.  Although much effort has been devoted to removing occlusions from face images, the varying shapes and textures of occlusions still challenge the robustness of current methods. As a result, current methods either rely on manual occlusion masks or only apply to specific occlusions. This paper proposes a novel face de-occlusion model based on face segmentation and 3D face reconstruction, which automatically removes all kinds of face occlusions with even blurred boundaries,e.g., hairs. The proposed model consists of a 3D face reconstruction module, a face segmentation module, and an image generation module. With the face prior and the occlusion mask predicted by the first two, respectively, the image generation module can faithfully recover the missing facial textures. To supervise the training, we further build a large occlusion dataset, with both manually labeled and synthetic occlusions. Qualitative and quantitative results demonstrate the effectiveness and robustness of the proposed method. 
\end{abstract}

\section{Introduction}
Thanks to deep Convolutional Neural Networks (CNNs), massive training data, and the widespread use of cameras, face-related techniques have successfully found many applications in our daily life, covering the fields of entertainment, media, art, security, etc. However, face images in the wild may be occluded by various objects, which leads to loss of information and undesirable noise, further resulting in degraded performance of algorithms for face analysis. To address this issue, researchers typically combine methods such as augmenting training data with synthesized occlusions~\cite{lv2017data,trigueros2018enhancing}, designing more sophisticated metrics and network structures~\cite{xia2015face,wan2017occlusion,song2019occlusion,ben2021face}, exploiting elaborated training strategies~\cite{he2016multiscale,wu2015robust,he2018dynamic}. Despite their effectiveness, most of them are task-specific and cannot be migrated to other face-related tasks. A relatively more general solution is to de-occlude the face image before passing it to the downstream tasks.   

\begin{figure}[!t]
    \centering
    \includegraphics[width=\linewidth]{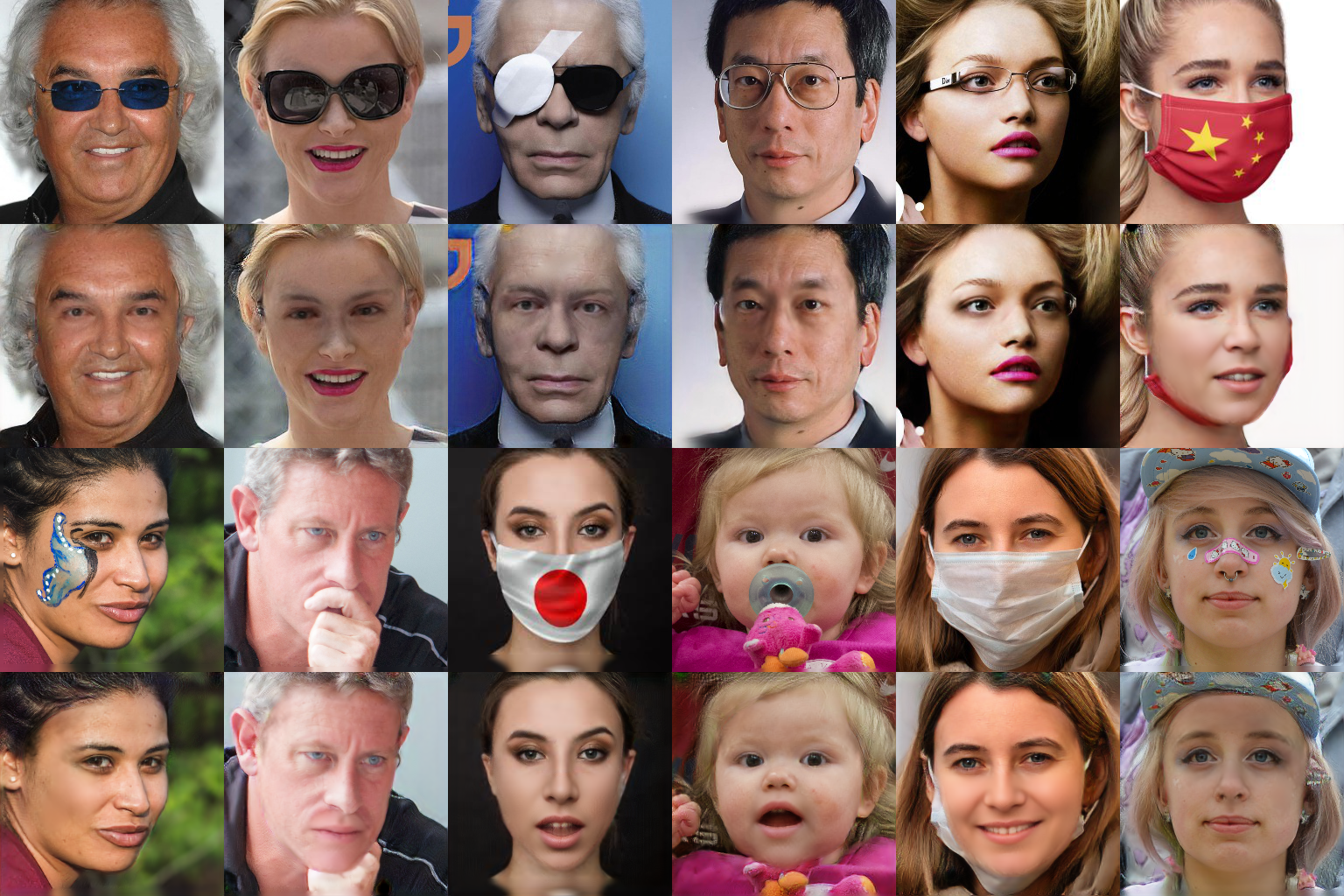}
    \caption{Results of the proposed method. The first row and the third row are input images, the second row and the last row show the de-occlusion results. Zoom in for a better view.}
    \label{fig:fig1}
\end{figure}

Before the widespread use of deep learning-based methods, the dominant approaches were based on matching and copying visible patches to the missing parts. A representative one is PatchMatch~\cite{barnes2009patchmatch}, which recursively searches the nearest neighbor textures to fill in the holes. Such copy-and-past-based methods work well in recovering simple low-frequency textures and obtain smooth results; however, they cannot recover high-level textures with complex structures, e.g., eyes, nose, mouth. ~\cite{turk1991eigenfaces,wu2004automatic,park2005glasses,wang2007reconstruction} leveraged Principal Component Analysis (PCA) to recover high-level semantic face textures. Nevertheless, such methods are based on the assumption that face images are in a linear space. Even for images acquired under constrained conditions, the results are blurry and far from satisfactory. As pointed out by ~\cite{hinton2006reducing}, the deep autoencoder is a non-linear generalization of the PCA. Therefore, it is reasonable to replace the PCA part of the above methods with an autoencoder. SSDA~\cite{xie2012image} proposed a deep sparse autoencoder to remove Gaussian noises and superimposed text from images, pioneered the use of deep neural networks in this area. Influenced by the great success of  Generative Adversarial Networks (GAN)~\cite{goodfellow2014generative} in image generation, Context Encoder~\cite{pathak2016context} introduced adversarial training to image inpainting for the first time. Its impressive performance led to the prosperity of GAN-based image inpainting methods. Since then, the combination of autoencoder and discriminator has been adopted as the basic model structure for image inpainting tasks~\cite{li2017generative, yu2018generative,liu2018image,yu2019free}. Despite the remarkable progress so far achieved, current methods mainly focus on filling the holes with visually plausible textures and neglect the detection of imperfectness. When applied to the face de-occlusion task, images must be accompanied by manually labeled masks. 
One strategy to address this problem is to train the model to detect the occlusions automatically. Due to the lack of well-labeled paired training data, existing methods mainly train on the synthetic occluded face images~\cite{zhao2017robust,cai2020semi,dong2020occlusion,yuan2019face,lee2020byeglassesgan,hu2020unsupervised}. Unfortunately, these methods either generate low-resolution results or only tackle specific occlusion types. We attribute this to the limited variety of synthetic occlusions. Real occlusions have different shapes, textures, and blurred boundaries, making it infeasible for synthesis-based methods to cover all possibilities of occlusion. In addition, collecting completely occlusion-free face images as ground truth is not easy because the forehead part of the face is often occluded by bangs, which is perhaps why none of the existing methods consider hair as a source of occlusion. 

Given the limitations of existing methods, this paper explores how to enable neural networks to remove arbitrary kinds of face occlusions. Our main contributions are summarized as follows:
\begin{itemize}
    \item We propose a novel face de-occlusion framework robust to all types of occlusions, regardless of their shapes and textures.
    \item We build a large occlusion dataset with extensive manually labeled occlusions from real face images. 
    \item The proposed method outperforms the state-of-the-art baselines quantitatively and qualitatively. 
\end{itemize}

\section{Related Work}
\subsection{Mask-Dependent Face Inpainting} 
Image inpainting aims to recover the missing textures of the input image. Recently, the autoencoder-GAN-based models have achieved impressive results. Benefiting from such structure, Context Encoder~\cite{pathak2016context} successfully predicts the $64\times 64$ missing part of $128\times 128$ images. To make the model further concentrate on the missing parts, ~\cite{yu2018generative, li2017generative} leverage two discriminators: a global discriminator, which takes the whole image as input, and a local discriminator, which only takes the small region around the missing part. However, such a design is only suitable for a single rectangular(or even square) hole and cannot be applied to images with irregular and arbitrarily distributed holes. To properly handle free-form masks, ~\cite{liu2018image} proposed a Partial Convolutional Layer, comprising a masked and re-normalized convolution operation followed by a mask-update step. ~\cite{yu2019free} proposed a Gated Convolution module, which learns a soft mask for the features in different layers. However, due to the ineffectiveness of CNNs in modeling long-range correlations between distant textures and the hole regions, the inpainting results often have boundary artifacts and look unreal.  ~\cite{yu2018generative} proposed a Contextual Attention module, which utilizes the image features as a convolution kernel, forcing the distant textures to interact with each other; this could be simply achieved by attention. ~\cite{zhang2019self} proposed a Self Attention module, which allows long-range dependency modeling of features; since then the attention mechanism has been widely applied in image generation tasks~\cite{cao2019gcnet,mejjati2018unsupervised,lei2020face,zheng2019pluralistic}. To make the generated image faithfully recover the facial topology, ~\cite{yang2020generative} leverages facial landmarks as a shape prior. ~\cite{zheng2019pluralistic} proposed a two-path probabilistic framework to generate pluralistic inpainting results. The common problem of the above methods is that they cannot detect the occluded parts automatically and rely on manually specified masks, limiting their usage scenarios.


\begin{figure*}[!ht]
    \centering
    \includegraphics[width=\linewidth]{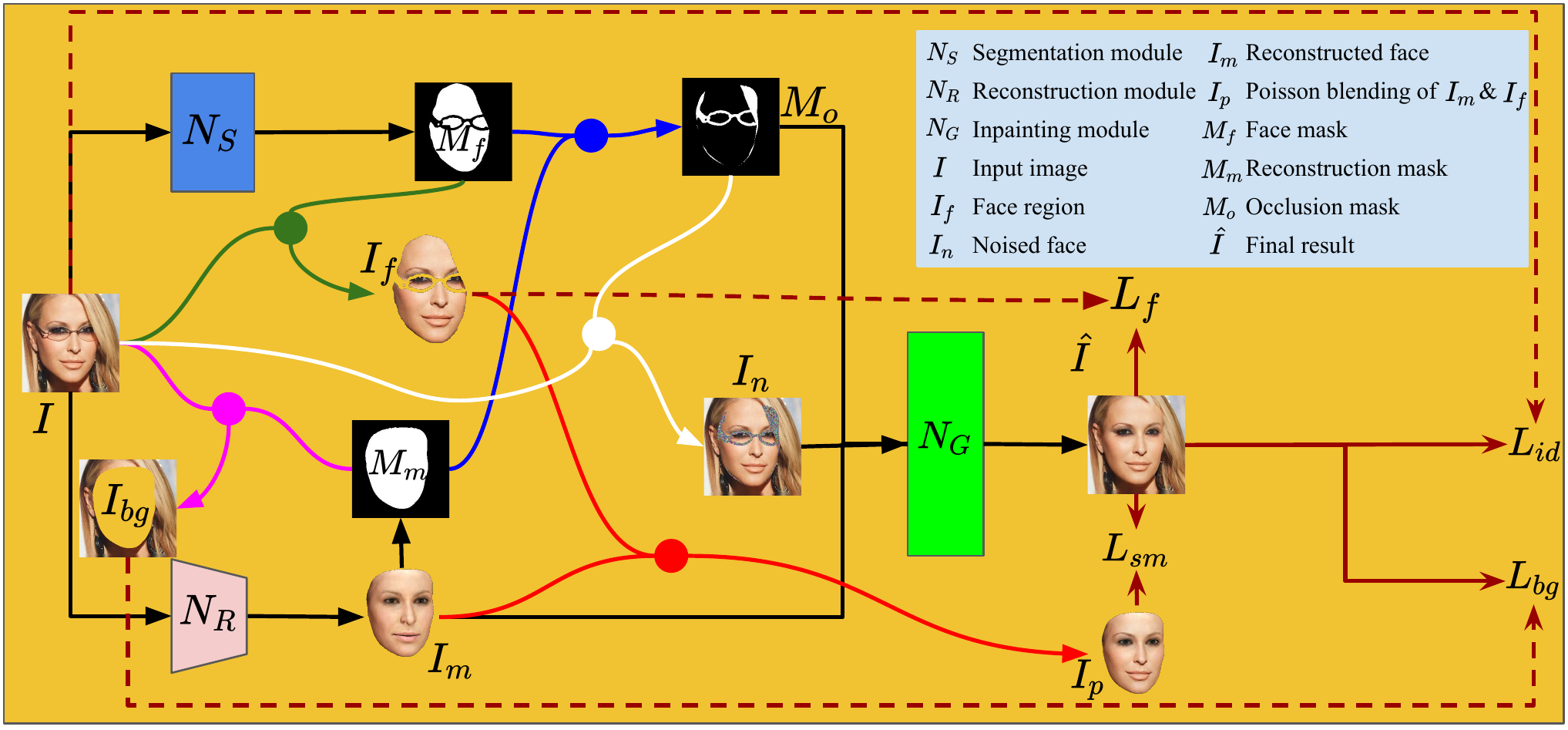}
    \caption{Overview of the proposed method, which does not include the discriminator part for space limitation}
    \label{fig:overview}
\end{figure*}

\subsection{Mask Free Face De-occlusion} 
Some methods are devoted to getting rid of masks' dependence and making the face de-occlusion process fully automated.  They are typically trained on face pairs with synthesized occlusions. ~\cite{zhao2017robust} proposed an LSTM-Autoencoder to gradually substitute the occlusions with facial textures. Benefiting from an elaborated forward pass, ~\cite{cai2020semi} proposed a semi-supervised face de-occlusion method, which is trained to predict the occlusion mask without the supervision of ground-truth. ~\cite{dong2020occlusion} proposed a two-stage GAN, where the first GAN aims to reconstruct the occlusion part solely, and the second GAN takes the result of the first GAN as a hint making the input image occlusion-free. Due to the limited variety of occlusions in the training set, the above methods can only generate images in $128\times 128$ resolution with noticeable artifacts, restricting their application to face recognition. Given the complexity of occlusion types, a relatively simple task would be to deal with only specific occlusion: ~\cite{lee2020byeglassesgan,hu2020unsupervised} for eyeglasses and ~\cite{din2020novel} for masks. Besides being limited by the occlusion variety, as they lack face topology information, the results are often far from satisfactory, especially on large-angled face images.

\subsection{3D-Guided Face De-occlusion}
The work closest to ours is FD3D~\cite{yuan2019face},  where we both use 3D Morphable Model~\cite{blanz1999morphable} (3DMM)-based reconstruction results as guidance. FD3D feeds the synthetic occluded face and the 3D reconstructed one to an adversarial auto-encoder and expects the output to be occlusion-free. Our method surpasses theirs in the following aspects: 1) They inpaint directly on the occluded face despite the challenge of converting various occlusion textures to face textures; in contrast, our method explicitly removes occlusions with predicted masks before inpainting, bypassing the disturbance of the occlusion textures. 2) FD3D entirely relies on supervised learning, which requires occlusion-free images to serve as the ground truth, while our method can be trained on initially occluded images. 3) They synthesize images with fixed occlusions, namely, cup, glasses, hand, mask, and scarf, which have 99 variations in total; by detecting the face mask instead of innumerable types of occlusions, our method can handle all types of occlusion.

\section{Proposed Method}
\subsection{Overview}
Figure~\ref{fig:overview} illustrates the overview of our method. Given that covering all possible occlusions is impossible, we consider the opposite direction. We employ a face segmentation module $N_S$ to predict the face mask $M_f$, which is much easier than predicting the mask of all kinds of occlusions. Then, we use the 3D reconstruction module $N_R$ to predict the prior 3DMM-based texture and the corresponding mask for the whole face, denoted as $I_m$ and $M_m$, respectively. The mask of the occlusions $M_o$ can then be calculated as:
\begin{equation}
\label{eq1}
    M_o = M_m - M_m\odot M_f,
\end{equation}
where $\odot$ denotes element-wise product. For face inpainting, we replace the occluded part of the input image $I$ with Gaussian noise to obtain $I_n$, which minimizes the disturbance of the occlusion textures. $I_n$, $I_{m}$, and $M_o$ are concatenated and fed into the inpainting module $N_G$, and the output $\hat{I}$ is expected to be occlusion-free. The following terms supervise the inpainting: 1) the background region $I_{bg}$, 2) the non-occluded face region $I_f$, 3) the Poisson blending result of $I_f$ and $I_m$, denoted as $I_p$. 

The overall framework contains three modules, $N_S$, $N_R$, and $N_G$, for face segmentation, 3D reconstruction, and image inpainting. We present the implementation details of each module separately in the following. 

\subsection{Face Segmentation}
\begin{figure}[t]
    \centering
    \includegraphics[width=\linewidth]{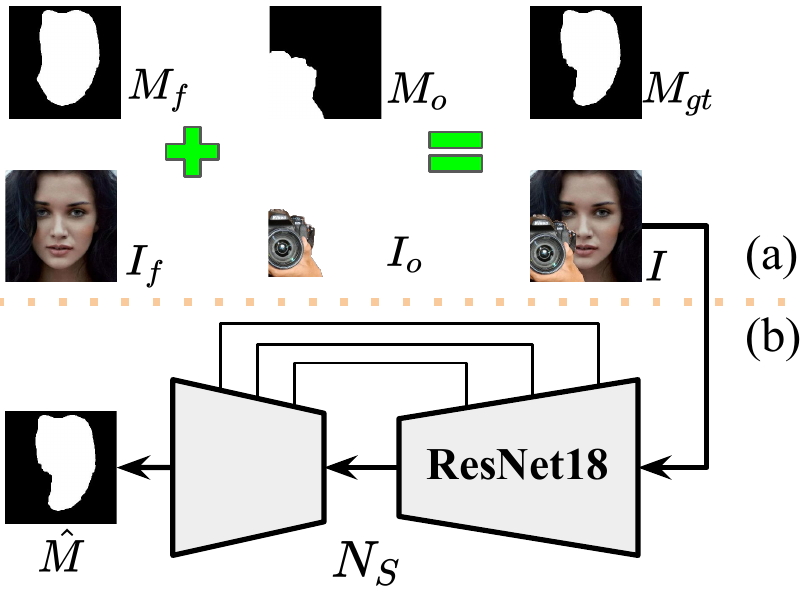}
    \caption{(a) Data augmentation method. $I_f$: face image, $M_f$: face mask of $I_f$, $I_o$: occlusion, $M_o$: occlusion mask, $I$: occluded image, $M_{gt}$: face mask of $I$. (b) Structure of the segmentation model $N_S$, $\hat{M}$: predicted face mask.}
    \label{fig:fig3}
\end{figure}
Extracting face regions from occluded face images is a simple, valuable, but never seriously addressed problem.  As a result, the recent 3D face reconstruction method~\cite{deng2019accurate} requires this functionality still needs to resort to a Naive Bayesian classifier. We present our solution to the problem in this section. 

\textbf{Data Preparation.}
We build our training data based on CelebAMask-HQ~\cite{CelebAMask-HQ}, which has 30,000 high-resolution face images, each with a segmentation mask of facial attributes.  Basically, given a segmentation mask of a face image, we can easily get the face region by gathering the labels of eyes, nose, lips, mouth, and skin. Unfortunately, the segmentation mask in the dataset does not consider the occlusions, e.g., hands, accessories, microphones, which are all misclassified as skin. In addition, the segmentation masks in the dataset do not distinguish between sunglasses and spectacles; however, in our task, we should not discard the skin and eyes under the transparent lenses of the spectacles. To solve the above problems, we manually picked out the occluded face images with incorrect segmentation masks and labeled more than 1700 occlusions from them. We also collected about 300 occlusions from Google to get a more balanced occlusion type (e.g., masks, scarfs). 
In addition, we collected more than 800 texture patches covering every type we can think of, which will substitute for the original occlusion textures during training to produce more occlusion variations. Those occlusions are utilized for data augmentation, as illustrated in Figure~\ref{fig:fig3} (a) and the following equation. 
\begin{equation}
\label{eq2}
\begin{aligned}
    &M_{gt} = M_f\odot(1-M_o), \\
    &I = I_f\odot(1-M_o) + I_o\odot M_o.  
\end{aligned}
\end{equation}

\textbf{Model Structure.} $N_S$ follows the classical U-Net structure~\cite{ronneberger2015u}, with ResNet18~\cite{he2016deep} as encoder, while the decoder is a stack of convolutional blocks corresponding to each stage of the encoder. Each decoder block takes features from its previous block and its symmetric encoder stage through skip-connection. Since the target mask is binary, a single channel output for the decoder suffices. 

\textbf{Losses.} The training is guided by the Dice loss~\cite{milletari2016v} and the Binary Cross-Entropy loss (BCE): 
\begin{equation}
    \label{eq3}
    L_{dice} = 1-\frac{2 \sum \hat{M}\odot M_{gt}}{\sum\hat{M} + \sum M_{gt}},
\end{equation}
\begin{equation}
\label{eq4}
\begin{aligned}
    L_{bce} =-&\frac{1}{WH} \sum \big( M_{gt}\odot\mathrm{log}\hat{M}+\\
    &(1-M_{gt})\odot\mathrm{log} (1-\hat{M})\big),
\end{aligned}
\end{equation}
where $W$ and $H$ represent the width and height of the mask, respectively. Further, we apply the ``online hard example mining" (OHEM)~\cite{shrivastava2016training} strategy to the BCE loss to make the training focus on hard examples, thus achieving more effective and efficient training.  

The proposed segmentation module achieves a precision of 0.981 and an IoU score of 0.954 on the validation dataset, which is acceptable considering the inaccurate edge region of the ground truth. By random checking the results, we surprisingly find that the predicted face region is often more accurate than the ground truth. We use this module to extract the face regions of the images in CelebA-HQ and FFHQ, which greatly facilitates the training of the other two modules, i.e., the reconstruction module $N_R$ and the inpainting module $N_G$.

\subsection{3D Face Reconstruction}
The reconstruction module is adapted from ~\cite{deng2019accurate}, the state-of-the-art 3D reconstruction method with a ResNet50~\cite{he2016deep} as its backbone. Given a face image, it predicts a vector $c\in \mathbb{R}^{239}$, containing 6 translation and rotation parameters; 144 shape and 80 texture coefficients of the 3DMM; and nine illumination coefficients of the Spherical Harmonics~\cite{ramamoorthi2001efficient,ramamoorthi2001signal} model. With vector $c$, the face is reconstructed and further rendered to the image through a differentiable renderer.


\begin{figure}[!t]
    \centering
    \includegraphics[width=\linewidth]{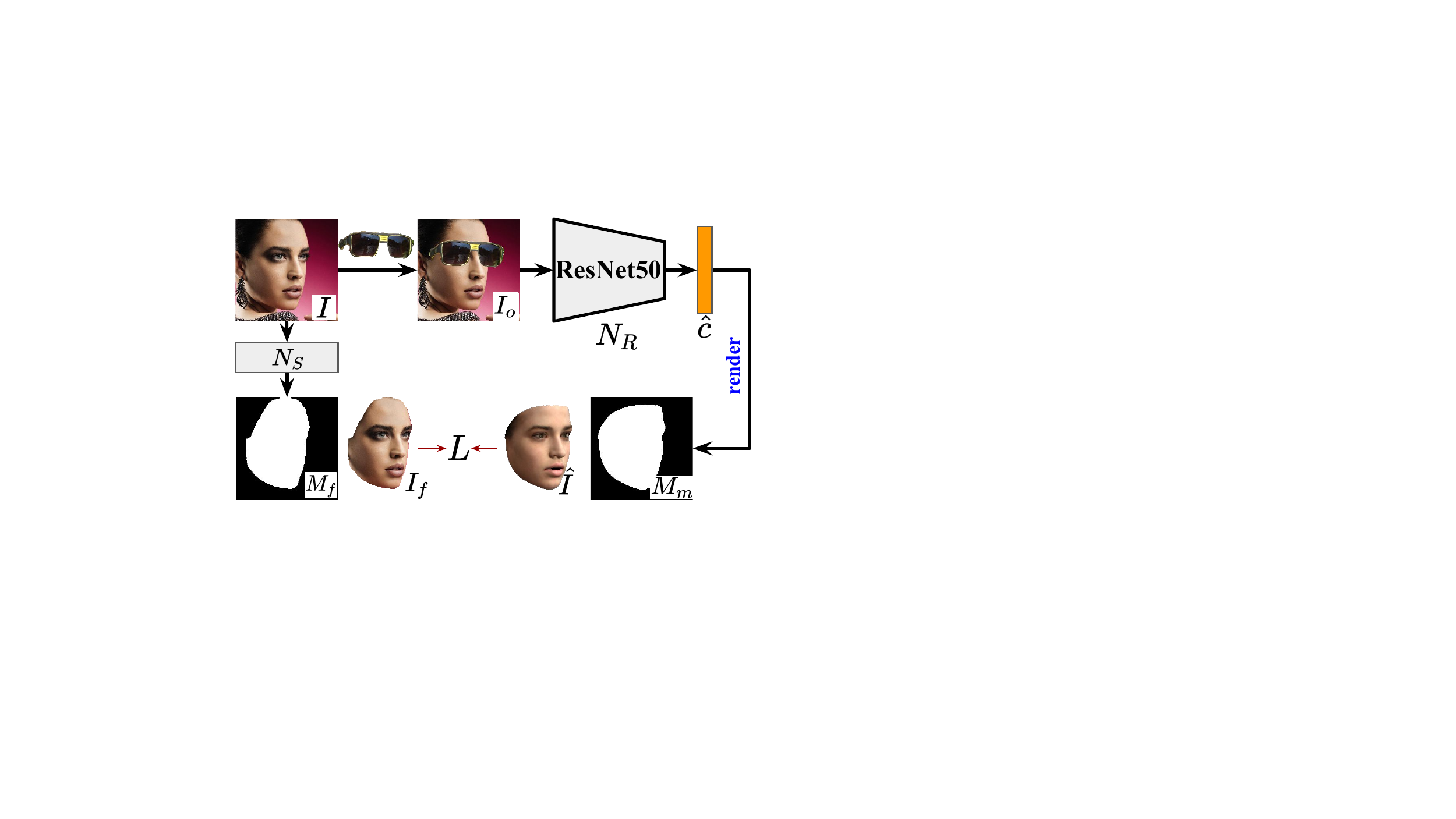}
    \caption{Overview of the 3D reconstruction module. $I$: original image, $I_o$: occluded image, $\hat{c}$: predicted 3D parameters, $\hat{I}$: reconstructed face, $M_m$: mask of $\hat{I}$, $M_f$: face mask of $I$, $I_f$: $I\odot M_f$}
    \label{fig:fig4}
\end{figure}

Although~\cite{deng2019accurate} performs well on occlusion-free faces and even demonstrates certain robustness to small occlusions, it still cannot handle severe occlusions, e.g., sunglasses and masks. As shown in the second row of Figure~\ref{fig:fig5}, sunglasses make the reconstructed face have dark circles. In the third column, they even change the color of the reconstructed skin. Therefore, in our usage scenario, i.e., face de-occlusion, it is necessary to retrain an occlusion-robust 3D face reconstruction module. Now we present, as shown Figure~\ref{fig:fig4}, our training strategy. 

\textbf{Training data.} The training is based on CelebA-HQ and FFHQ. Firstly, we use the pre-trained model of~\cite{deng2019accurate} to predict the reconstruction parameters of the training data, denoted as $c_{gt}$. Then, we filter out all the images with sunglasses since their corresponding $c_{gt}$ risk to be inaccurate. Next, we leverage the face segmentation module $N_S$ to detect the face masks of the training data, denoted as $M_f$, for calculating the face-related losses. At last, as with the previous section, we randomly superimpose heavy occlusions to $I$ during training to get occluded input images for $N_R$. 

\textbf{Losses.} Since we already have $c_{gt}$, the most straightforward option is to use it as supervision; the loss is as:
\begin{equation}
    \label{eq5}
    L_{coef} = \frac{1}{N}\|\hat{c} - c_{gt}\|_1,
\end{equation}
where $\hat{c}$ is the predicted coefficient, and $N$ is its dimension.

Training solely with Equation~\ref{eq5} will result in inaccurate and non-discriminative results. Since it equally optimizes all the coefficients. However, different parts of $\hat{c}$ obviously have different impacts on the reconstruction result, e.g., the poses play a more critical role than the illumination coefficients. Therefore, we also leverage the pixel-wise and the perceptual~\cite{johnson2016perceptual} losses to reduce the discrepancy between the reconstructed face $\hat{I}$ and the real face $I_f$:   

\begin{figure}[!t]
    \centering
    \includegraphics[width=\linewidth]{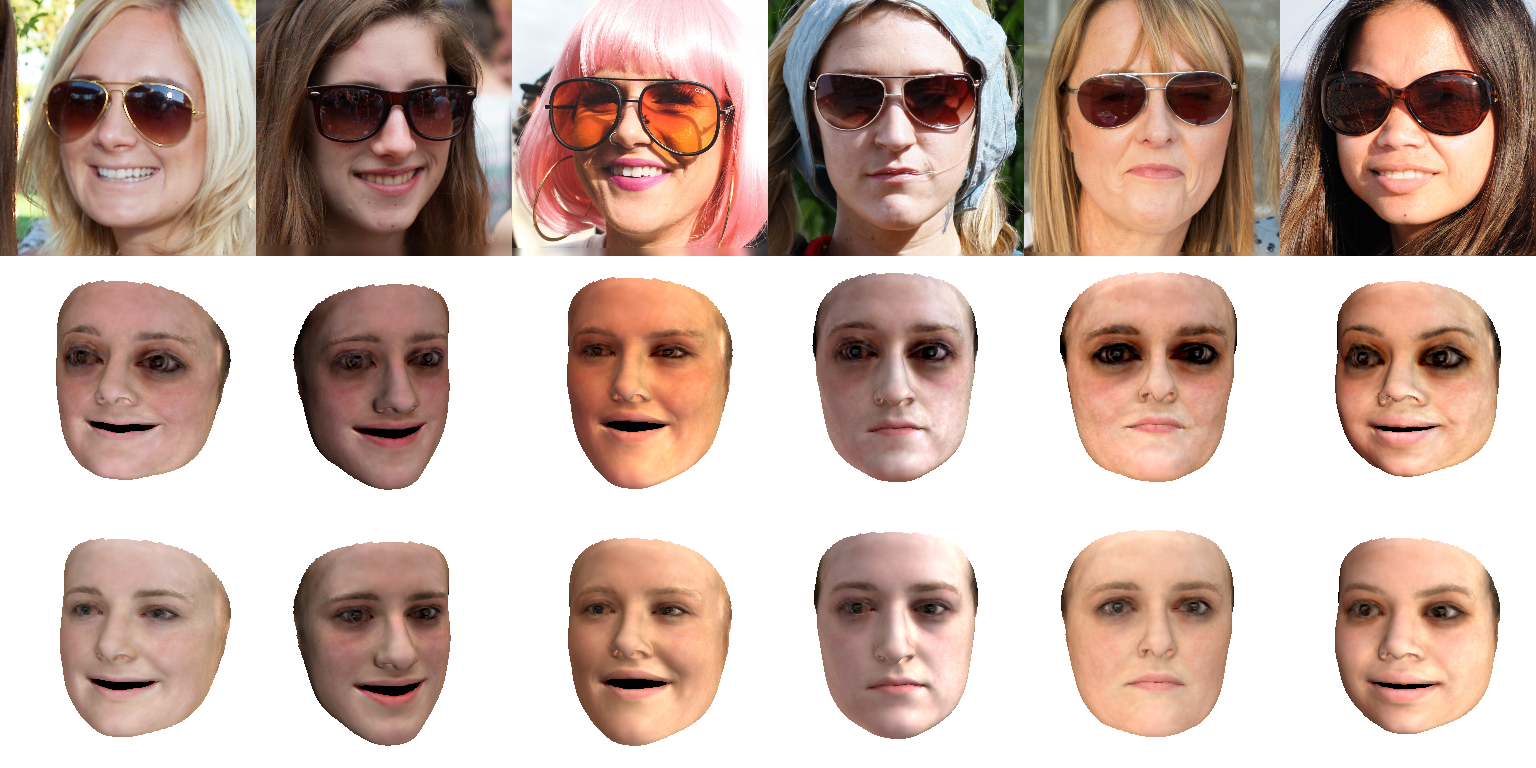}
    \caption{The first row shows the input images; the second row and the last row are reconstructed faces by~\cite{deng2019accurate} and ours, respectively.}
    \label{fig:fig5}
\end{figure}

\begin{equation}
\label{eq6}
    L_{pix} = \frac{1}{\sum M}\|\hat{I}\odot M - I_f\odot M \|_2,
\end{equation}
\begin{equation}
    \label{eqid}
    L_{id} = 1- \frac{\mathcal{F}(\hat{I})^T\cdot \mathcal{F}(I_f)}{\|\mathcal{F}(\hat{I})\|\cdot \|\mathcal{F}(I)\|},
\end{equation}
where $M$ is the overlap of the original face mask $M_f$ and the reconstruction mask $M_m$;  $\mathcal{F}(\cdot)$ denotes the feature embedding function of a pre-trained face recognition model, and we use Arcface~\cite{deng2019arcface} here. We only extract the feature of $I_f$, rather than the entire image $I$, to avoid the noises introduced by the non-facial patterns.  

To accelerate the training, we also use the landmark loss. Thanks to $c_{gt}$, we have ground truth 3D facial landmarks $\mathbf{q}$, thus eliminating the need for a 2D facial landmark detector as required in other methods. The loss is calculated as :
\begin{equation}
    \label{eq7}
    L_{ldmk} = \frac{1}{n_{pt}}\sum_{i=1}^{n_{pt}} \omega_n \|\hat{\mathbf{q}}_i-\mathbf{q}_i\|^2,
\end{equation}
where $n_{pt}$ denotes the number of landmarks, $\hat{\mathbf{q}}_i$ denotes the predicted 3D coordinates of the $i$-th facial landmark. The weights $\omega_n$ are set to 20 for the nose and inner mouth points and 1 for others. 

As $c_{gt}$ provides sufficient regularity, we discard the complex regularization terms in~\cite{deng2019accurate}. The overall loss is the weighted sum of the above losses: 
\begin{equation}
    \label{eq8}
    L = L_{coef} + \lambda_{pix} L_{pix} + \lambda_{id} L_{id} + \lambda_{ldmk} L_{ldmk},
\end{equation}
where $\lambda_{pix}=1.92$, $\lambda_{id}=0.2$, $\lambda_{ldmk}=1.6e^{-3}$. 

Figure~\ref{fig:fig5} compares our results with those of ~\cite{deng2019accurate}, demonstrating the effectiveness of the proposed training strategy in improving the occlusion-robustness of the model.  A quantitative comparison is also performed: we use ~\cite{deng2019accurate} and our model to reconstruct 1000 face images with synthetic occlusions, the $L_{pix}$ of our model is 0.153, which is much less than 0.177 of~\cite{deng2019accurate}.

\subsection{Face Inpainting}
With the $N_S$ and $N_R$ described above, we get the following information of an occluded face image: face mask $M_f$, face $I_f$, reconstructed face $I_m$, reconstructed face mask $M_m$. Based on $M_f$ and $M_m$, Equation~\ref{eq1} further calculates the occlusion mask $M_o$. The goal of this section is to restore the missing textures indicated by $M_o$. 

\begin{figure}[t]
    \centering
    \includegraphics[width=\linewidth]{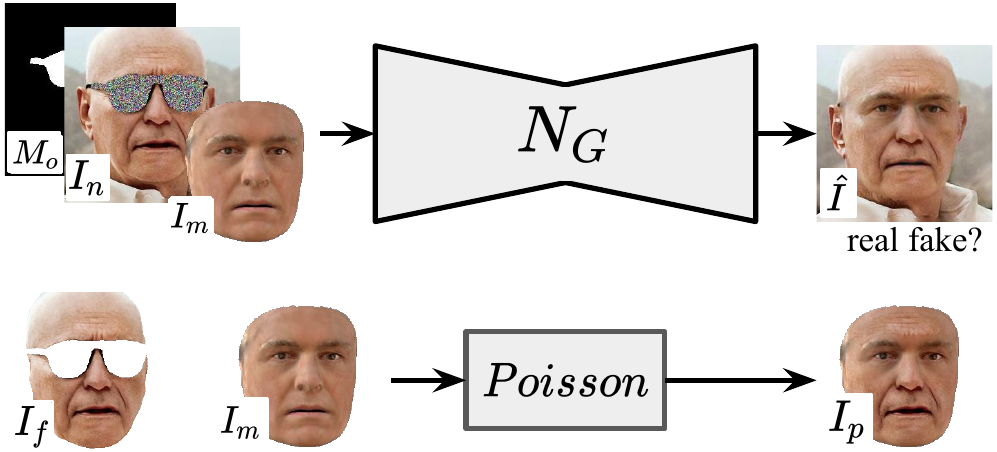}
    \caption{Overview of the inpainting module. }
    \label{fig:fig6}
\end{figure}

The generator $N_G$ is constructed from stacked gated residual blocks~\cite{yu2019free} and follows the classical encoder-decoder structure; a self-attention~\cite{zhang2018self} module is applied to the bottleneck features. As shown in Figure~\ref{fig:fig6}, the input is composed of three parts: the occlusion mask $M_o$, the face image with noised occlusion $I_n$, and the reconstructed face $I_m$. The output $\hat{I}$ is the de-occluded image. A VGG~\cite{simonyan2014very} shaped discriminator is further employed to increase the photorealism to $\hat{I}$. 

We randomly superimpose occlusions to the training data and train $N_G$ to recover the missing textures. One issue to note is that the input image itself always contains occlusions, such as bangs, and we do not have ground truth for these parts. So we use Poisson Blending~\cite{perez2003poisson} to seamlessly merge $I_f$ and $I_m$ and get $I_p$, which provides weak supervision for the inpainting task. The training is guided by the following losses:

\textbf{Pixel-wise face loss.}
\begin{equation}
    \label{eq10}
    L_{pix} = \frac{1}{\sum M} \|\hat{I}\odot M -I_f \odot M\|_1,
\end{equation}
where $M=M_m\odot M_f$, limiting the loss to be calculated on the face region solely. This loss serves for recovering textures occluded by the synthetic occlusion and cannot recover the initially occluded textures.

\textbf{SSIM loss.} We use $I_p$ to guide the generation of initially occluded textures. Although the Poisson blending result is visually pleasing, it changes the color of $I_f$, so we cannot simply apply $L_1$ or $L_2$ loss. Instead, we leverage the Structural Similarity loss (SSIM)~\cite{wang2004image}, emphasizing the structural level discrepancy: 
\begin{equation}
    \label{eq11}
    L_{sm} = \frac{-1}{\sum{\bar{M}_m}} SSIM(\hat{I}\odot \Bar{M}_m, I_p\odot\bar{M}_m),
\end{equation}
where $SSIM$ stands for the similarity mapping function (see Appendix for details); $\bar{M}_m$ is the eroded $M_m$, eliminating the edge effects in the similarity map.  To further mitigate the impact of inaccurate color of $I_p$ and make the loss focus on the missing textures, we apply the OHEM again as in Equation~\ref{eq4}.

\textbf{Background loss.} Areas other than the face should remain unchanged:
\begin{equation}
    \label{12}
    L_{bg} = \frac{1}{\sum M_{bg}} \|\hat{I}\odot M_{bg}-I\odot M_{bg}\|,
\end{equation}
where $M_{bg}$ denotes the background mask, calculated by $1-M_m$. We also erode $M_bg$ to alleviate the edge effects. 

\textbf{Identity loss.} The generated image should have the same identity as the original image, so we also leverage Equation~\ref{eqid} to add feature level identity constraints.

\textbf{TV loss.} To penalize the noises in $\hat{I}$, we adopt the total variation loss~\cite{mahendran2015understanding}. 
\begin{equation}
    \label{eq13}
    L_{tv} = \frac{1}{WHC}\|\nabla_x\hat{I}\|^2 + \|\nabla_y\hat{I}\|^2 ,
\end{equation}
where $W$, $H$, $C$ are the width, height, and the number of channels, respectively, $\nabla\_$ calculates the image gradient along a direction.  

\textbf{Adversarial loss.} To further make $\hat{I}$ photorealistic, we employ the adversarial loss: 

\begin{equation}
    \label{eq14}
    L_{adv} = -\mathbb{E}_{\hat{I}}[\mathrm{log}(D(\hat{I}))],
\end{equation}
where $D(\cdot)$ denotes the output of the discriminator, which is the term to be maximized by the generator. 

The global objective function of the generator can be summarized as follows: 
\begin{equation}
    \label{eq15}
    \begin{aligned}
        L =& \lambda_{pix}L_{pix} + \lambda_{sm}L_{sm} + \lambda_{bg}L_{bg} +\\ &\lambda_{id}L_{id} + \lambda_{tv}L_{tv} + \lambda_{adv}L_{adv},
    \end{aligned}
\end{equation}
where the weights are empirically set to $\lambda_{pix}=10$, $\lambda_{sm}=5$, $\lambda_{bg}=5$, $\lambda_{id}=0.2$, $\lambda_{tv}=0.1$, $\lambda_{adv}=0.01$.

\textbf{Discriminator loss.} We use the BCE loss to train the discriminator, aiming to distinguish $\hat{I}$ from real images: 
\begin{equation}
    \label{eq16}
    L_{adv} = \mathbb{E}_I[\mathrm{log}(D(I))] + \mathbb{E}_{\hat{I}}[\mathrm{log}(1-D(\hat{I}))],
\end{equation}
where $I$ is the real image with less occlusion, selected from CelebA-HQ and FFHQ according to the overlap rate of its corresponding $M_m$ and $M_f$. 

\section{Experiments}
The proposed method can effectively remove the occlusions from face images. To demonstrate the power of our model, we qualitatively compare our results with state-of-the-art methods, including those for face de-occlusion and for image inpainting. Quantitative experiments are conducted in two aspects as well, namely reconstruction ability and identity recovery ability. 

\begin{figure*}[t]
    \centering
    \includegraphics[width=0.8\linewidth]{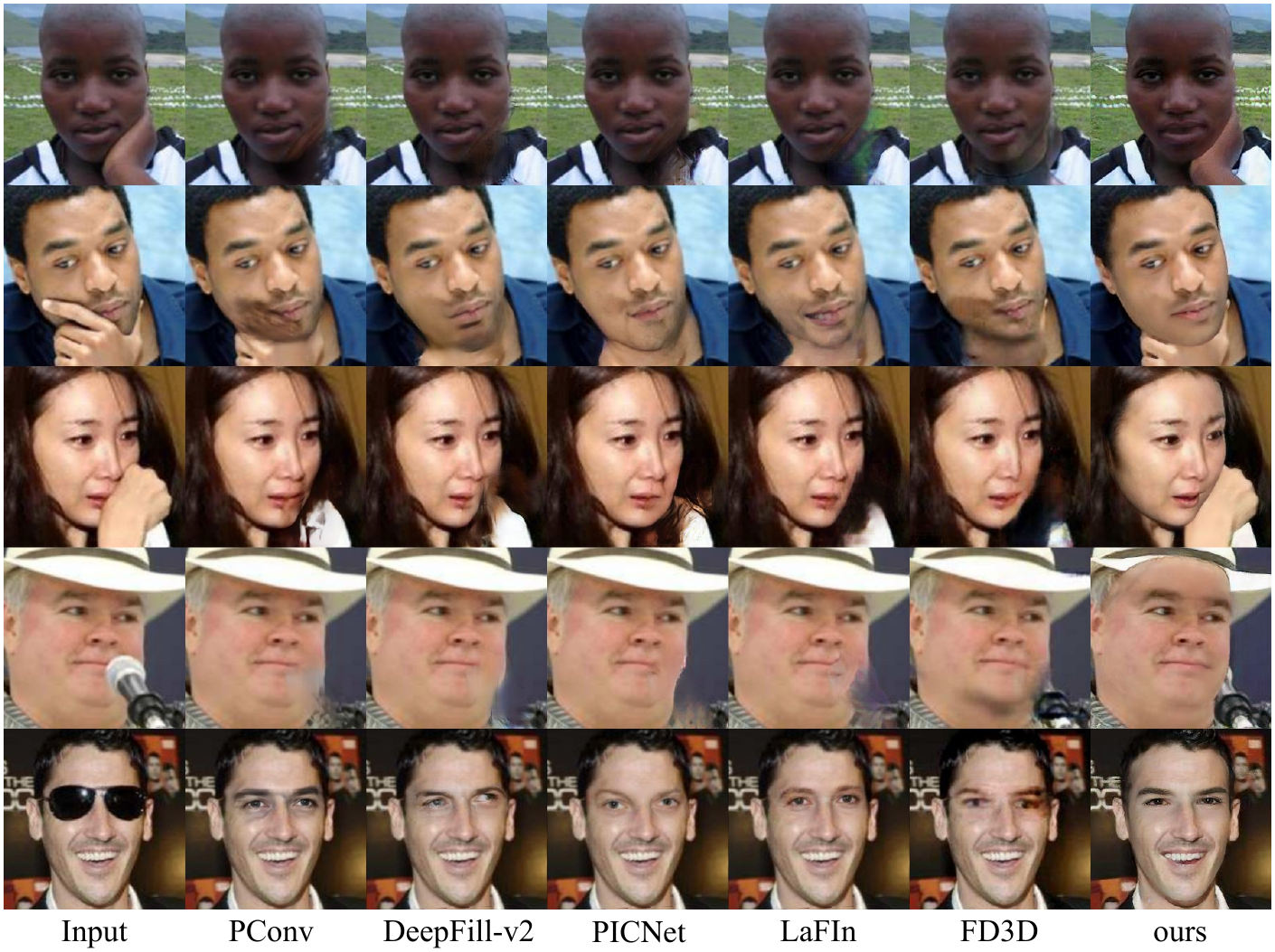}
    \caption{De-occlusion results compared with state-of-the-art methods. Input images are aligned and cropped according to the rules of each method, and we re-paste the outputs back to facilitate comparison.}
        \label{fig:abcde}
\end{figure*}

\subsection{Implementation Details}
We follow an incremental training strategy: 1) We train the segmentation module $N_S$. 2) we train the reconstruction module $N_R$ based on $N_S$. 3) We train the inpainting module $N_G$ based on the result of the first two. The training is performed on two public datasets, CelebAMask-HQ~\cite{CelebAMask-HQ} and FFHQ~\cite{karras2017progressive}, and our manually labeled/collected occlusions and textures.  All the images are aligned with facial landmarks detected by~\cite{bulat2017far} and cropped to $256\times 256$.  

The training of $N_S$ consists of two rounds: In the first round, we train on 300,000 images of CelebAMask-HQ with our manually labeled occlusions to obtain a coarse model. In the second round, we apply the coarse model to both CelebAMask-HQ and FFHQ~\cite{karras2019style}, select 500 hard examples from the results, relabel and add them to the occlusion dataset to retrain a more accurate model. Each round is trained for 30 epochs with a batch size of 16 and a learning rate of $1e^{-4}$. The training takes about two hours on two Nvidia GTX 1080 GPUs. The 3D reconstruction module and the face inpainting modules are trained on 100,000 images of CelebAMask-HQ and FFHQ, with an initial learning rate of $1e^{-4}$. For the reconstruction module $N_R$, we train 50 epochs with a batch size of 16, and for the inpainting module $N_G$, we train 80 epochs with a batch size of 4. The learning rates of both modules are dropped halfway through the training with ratios of 0.2 and 0.1, respectively. All three modules are optimized using Adam~\cite{kingma2014adam} with a weight decay of 0. Betas are set to $[0.9, 0.999]$ for $N_S$ and $N_R$, $[0.5, 0.999]$ for $N_G$. It takes about 80 hours to train $N_R$ and 40 hours to train $N_G$ on two Titan X Pascal GPUs.  

\subsection{Qualitative Results}
The method closest to ours is FD3D~\cite{yuan2019face}; we both leverage 3D face reconstruction for face de-occlusion. Unfortunately, they do not release their code, so we use images from their paper to conduct the qualitative evaluation.  Figure~\ref{fig:abcde} compares our method with FD3D and several other publicly available image inpainting methods, including DeepFill-v2~\cite{yu2019free}, PConv~\cite{liu2018image}, PICNet~\cite{zheng2021pluralistic}, and LaFIn~\cite{yang2020generative}. The FD3D results are taken from their paper, while the rest are generated from their official implementations with models pre-trained on face images. We provide those inpainting methods with manually labeled occlusion masks for a fair comparison. One issue to note is that when FD3D compares with other methods in their paper, the images are not pre-aligned according to the rules of the corresponding method, resulting in questionable inferior results.  

As shown in Figure~\ref{fig:abcde}, our method generally outperforms FD3D and other image inpainting methods, especially on face images with large yaw angles. We argue that the image inpainting methods are purely based on statistical learning, thus highly dependent on the training data distribution. When a few face images with large yaw angles occur, these methods record dramatic performance degradation. Although FD3D does not require a manual mask for the de-occlusion, they apparently can only handle limited types of occlusions: the bangs in the third row and the hat in the fourth row are not identified as occlusions. In addition, FD3D produces low-quality face images with blurred boundaries and unreal textures, as evidenced in the images from the second to the last row. We believe this is due to the following reasons: 1) they do not explicitly detect the occlusions and mask them out, causing the generator to only cope with the occlusions it has ever seen while failing to de-occlude arbitrary occlusions; 2) they use a very coarse 3D reconstruction method, which cannot provide correct and effective face prior to the generator.

\subsection{Quantitative Results}
This section evaluates the proposed method in two aspects: the reconstruction ability and the identity recovery ability. We synthesize 1000 sunglasses-occluded face images and use DeepFill-v2, PICNet, LaFIn, and our method to recover the initial images. Since the image inpainting methods only synthesize the missing region, the reconstruction ability is evaluated only in that region, with the following metrics: L1 loss, SSIM~\cite{wang2004image} score, and PSNR score. The identity recovery ability is evaluated by the cosine similarity of the features extracted by the pre-trained ArcFace~\cite{deng2019arcface}. Results are reported in Table~\ref{tab:my_label}. 

\begin{table}[t]
    \centering
    \normalsize
    \begin{tabular}{c c c c c}
    \hline
    Method & $L_{1}$$\downarrow$ & SSIM $\uparrow$ & PSNR$\uparrow$ & ID$\uparrow$\\
    \hline
    \hline
    LaFIn& 0.067 & 0.607 & 27.944 &  0.639 \\
    PICNet&0.0654 &0.604 & 28.137 & 0.615 \\
    DeepFill & 0.065 & 0.604 & 28.105 & 0.659 \\
    \hline
    w/o SSIM & 0.065 & 0.619 & 28.491 & 0.665 \\
    \textbf{Ours} & \textbf{0.062} & \textbf{0.623} & \textbf{28.902} & \textbf{0.690} \\
    \hline
    \end{tabular}
    \caption{Comparison of the proposed method with state-of-the-art image inpainting methods.}
    \label{tab:my_label}
\end{table}

As can be seen, our method outperforms the state-of-the-art image inpainting methods across all listed metrics. It performs particularly well in recovering the identity, much exceeding its comparators. We attribute this mainly to the occlusion-robust face reconstruction module. 

\subsection{Ablation Study}
The three modules in our framework are interdependent, making it impossible to remove one for ablation study. Instead, we mainly focus on the SSIM loss of the inpainting module, as the reconstruction and identity losses are conventional practices.

As Table~\ref{tab:my_label} shows, without the SSIM loss, the model degrades in all aspects of metrics (nevertheless, benefiting from the 3D prior, it still outperforms others). Moreover, the initially occluded textures are now only supervised by the adversarial loss; thus, the results are not guaranteed to be occlusion-free (see Appendix for examples).

\subsection{Discussion}
\begin{figure}[!t]
    \centering
    \includegraphics[width=\linewidth]{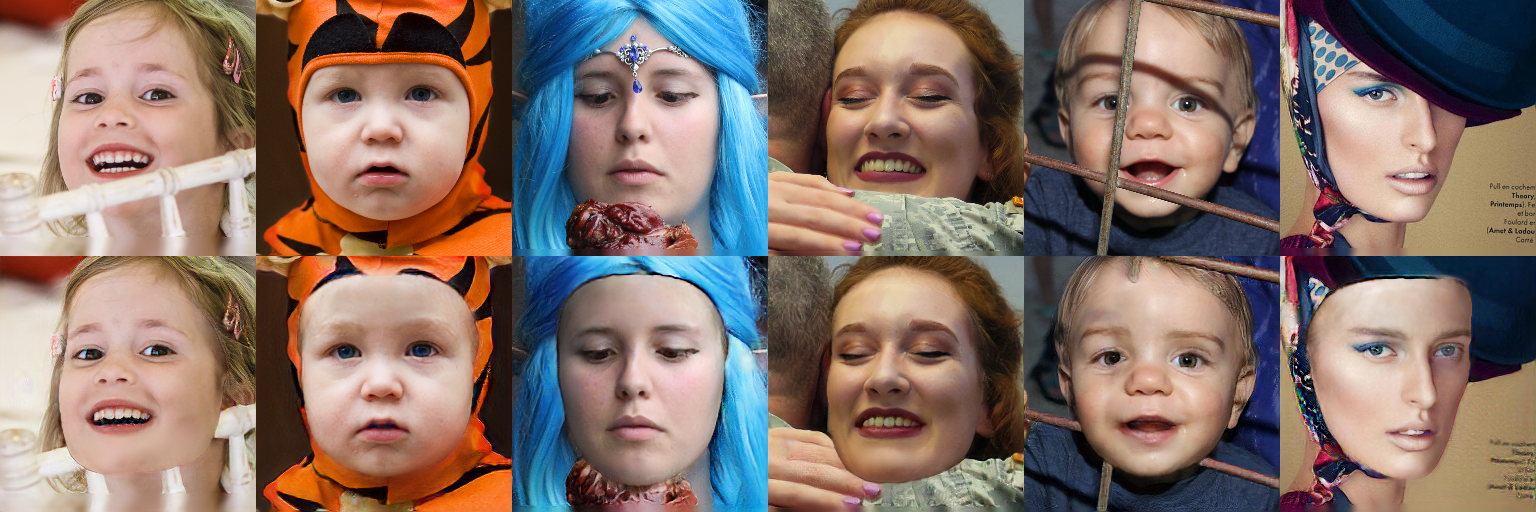}
    \caption{Limitation of the proposed method.}
    \label{fig:limit}
\end{figure}

\textbf{Limitation.}The proposed method focuses on removing the occlusions within the 3D reconstructed face mask. For occlusions spanning the face and the background, it can only rigidly remove the parts above the face without producing a smooth transition, creating an unrealistic scene where the face seems to hover above the occlusions, as Figure~\ref{fig:limit} shows. We have tried to create a gap between the face area and the background through mask erosion, and the gap is solely supervised by the adversarial loss during training.  This trick alleviates the boundary effect to some extent but still cannot cope with large-sized occlusions. 

\begin{figure}[t]
    \centering
    \includegraphics[width=0.8\linewidth]{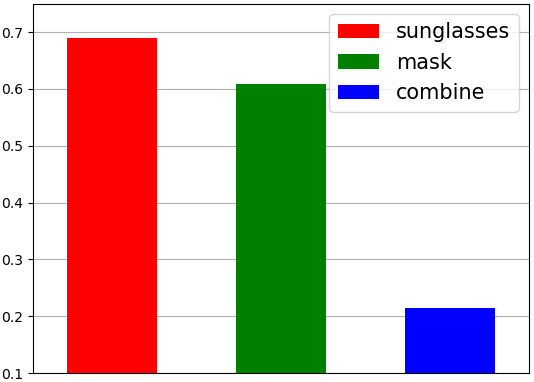}
    \caption{Identity recovery ability for different types of occlusion.}
    \label{fig:diff_occ}
\end{figure}

\textbf{Social impacts.}  Our method helps pre-process face images to avoid the negative impact of occlusions such as hair, glasses, hands, etc., on downstream tasks (e.g., fine-grained 3D face reconstruction, face recognition). Despite the benefits it brings, it also risks violating human privacy. To quantify this risk, we further conduct experiments to analyze the identity recovery ability across different types of occlusions. Figure~\ref{fig:diff_occ} shows that the proposed method performs well on sunglasses-occluded faces; the performance decreases for the mask-type occlusion; and when combining the mask with sunglasses, the method can no longer recover the face identity. The above observation proves that the method is controllable for privacy violations. People's identities can be safely protected when they simultaneously wear sunglasses and masks.

\section{Conclusion}
This work proposes a segmentation-3D reconstruction-guided facial image de-occlusion method that automatically removes all types of occlusions. We first analyze the limitations of current face de-occlusion methods: they either require a manually labeled mask or can only handle a limited number of occlusion types, which mainly stems from the vast diversity of possible occlusions. Our key innovation is bypassing the segmentation of infinite occlusions and instead segmenting the face regions, which is much easier. We manually labeled a large face occlusion dataset, based on which we trained a face segmentation module $N_S$ and an occlusion-robust 3D reconstruction module $N_G$. Given an occluded face image, $N_S$ and $N_G$ work collaboratively to mask the occlusions and provide beneficial priors to guide the subsequent face inpainting module. Qualitative and quantitative evaluations demonstrate the superiority of the proposed method.



{\small
\bibliographystyle{ieeetr}
\bibliography{egbib}
}

\end{document}